\RequirePackage{fix-cm} 
\documentclass[smallextended,11pt]{svjour3}  

\usepackage{amssymb}  
\usepackage{amsmath}
\usepackage{amsfonts}
\usepackage{booktabs}
\usepackage{makecell}
\usepackage{cite}
\usepackage{latexsym}
\usepackage{mathptmx} 
\usepackage{bm}      
\usepackage{bbm}      
\usepackage{graphicx} 
\usepackage{subfigure}
\usepackage{graphics} 
\usepackage{color}    
\usepackage{epsfig}   
\usepackage{subfigure}
\usepackage{epstopdf} 
\usepackage{algorithm}
\usepackage{algorithmicx}
\usepackage{mathrsfs} 
\usepackage{wasysym}  
\usepackage[colorlinks,linkcolor=blue,citecolor=blue]{hyperref}
\usepackage{float}    
\usepackage{multirow} 
\usepackage{comment}  
\usepackage{overpic}  
\usepackage{psfrag}   
\usepackage{rotating} 
\usepackage{stmaryrd} 
\usepackage{verbatim} 
\usepackage{ifthen}
\usepackage{pifont}   
\usepackage{url}
\usepackage{hyperref}
\usepackage{hyperref}
\usepackage{amsmath}

\usepackage[misc]{ifsym}

\usepackage[hang]{footmisc}

\usepackage{stfloats}

\hypersetup{hidelinks,
	colorlinks=true,
	allcolors=black,
	pdfstartview=Fit,
	breaklinks=true}

\usepackage{picinpar}   
\usepackage{graphicx}

\makeatletter
\long\def\figwindownonum[#1,#2,#3,#4] {
	\begin{window}[#1,#2,{#3},{\centering#4\par}] }
	\def\endfigwindownonum{\end{window}}
\makeatother

\smartqed             
\usepackage{enumitem} 

\newcounter{num}     
\numberwithin{equation}{section}

\numberwithin{theorem}{section}

\numberwithin{lemma}{section}

\numberwithin{corollary}{section}    

\begin{document}
\begin{sloppypar}
\title{BARS: A Benchmark for Airport Runway Segmentation}

\author{Wenhui Chen \textsuperscript{1}  \and
        Zhijiang Zhang \textsuperscript{1,~\Letter}  \and
        Liang Yu \textsuperscript{2}  \and
        Yichun Tai \textsuperscript{1}
        }

\date{}

\institute{
\begin{itemize}
      \item[] Wenhui Chen \\
            \email{wenhuichen@shu.edu.cn}\\
     \item[] {\Letter} Zhijiang Zhang \\
            \email{zjzhang@staff.shu.edu.cn}\\
    \item[] Liang Yu \\
        \email{yuliang1@comac.cc}\\
    \item[] Yichun Tai \\
            \email{taiyc@shu.edu.cn}
      \at
     \item[] {1} School of Communication and Information Engineering, Shanghai University, PR China
     \at
     \item[] {2} Shanghai Aircraft Design and Research Institute, PR China
     \end{itemize}
}

\maketitle
\begin{abstract}
Airport runway segmentation can effectively reduce the accident rate during the landing phase, which has the largest risk of flight accidents. With the rapid development of deep learning (DL), related methods achieve good performance on segmentation tasks and can be well adapted to complex scenes. However, the lack of large-scale, publicly available datasets in this field makes the development of methods based on DL difficult. Therefore, we propose a benchmark for airport runway segmentation, named BARS. Additionally, a semiautomatic annotation pipeline is designed to reduce the annotation workload. BARS has the largest dataset with the richest categories and the only instance annotation in the field. The dataset, which was collected using the X-Plane simulation platform, contains 10,256 images and 30,201 instances with three categories. We evaluate eleven representative instance segmentation methods on BARS and analyze their performance. Based on the characteristic of an airport runway with a regular shape, we propose a plug-and-play smoothing postprocessing module (SPM) and a contour point constraint loss (CPCL) function to smooth segmentation results for mask-based and contour-based methods, respectively. Furthermore, a novel evaluation metric named average smoothness (AS) is developed to measure smoothness. The experiments show that existing instance segmentation methods can achieve prediction results with good performance on BARS. SPM and CPCL can effectively enhance the AS metric while modestly improving accuracy. Our work will be available at \href{https://github.com/c-wenhui/BARS}{https://github.com/c-wenhui/BARS}.


\end{abstract}

\section{Introduction}

Aircraft flight phases include departure, cruising, and landing. Compared to other phases, the landing phase has the largest risk of flight accidents, as it is the most difficult phase to operate in. A successful landing requires maintaining the proper glide angle and descent speed, as well as ensuring that the aircraft's flight path is aligned with the runway centerline and crosses the intended landing point on the runway. Therefore, reducing pilot workload and improving safety during the landing phase are vital goals for the aviation industry. Existing landing phase navigation systems include the instrument landing system and ground-based augmentation system. However, the deployment cost of such systems is high. In recent years, the visual navigation system has emerged as a new development in this field due to its low cost.
Many studies \cite{tu2020airport,wang2022benchmark,kugler2019vision,krammer2020testing,krammer2021concept,Visual2022} are aimed at achieving automatic landing with the help of computer version technology. As an important part of the visual navigation system, airport runway segmentation classifies the runway markings at the pixel level, resulting in segmentation results that can indicate whether the aircraft is aligned with the runway centerline and whether the current glide angle is reasonable, which helps pilots better perceive the runway position, enabling automatic landing and improving safety during the landing phase.

Existing solutions to airport runway segmentation are mainly implemented by identifying runway characteristics such as textures \cite{zhang2020runway} and line segments \cite{tang2015novel,zhang2022runway}.
However, such traditional image processing methods provide limited categories and cannot distinguish instances in the same category. Moreover, these methods are difficult to adapt to complex scenes, as some unrelated objects with similar shapes or structures may decrease accuracy. Segmentation methods \cite{tong2021sat,sun2022instance,zhang2022lightweight,cang2019research} based on deep learning (DL) provide good generalization and performance but rely on related datasets. A few studies \cite{aytekin2013texture,akbar2019runway,men2020airport,wangsemantic} have proposed datasets for airport runway segmentation. However, the datasets in \cite{aytekin2013texture,akbar2019runway,men2020airport} are remote sensing image datasets taken from the Earth view, which are not applicable to aircraft landing phase scenes. Additionally, they all have a small quantity of data and are not publicly available. Due to the lack of relevant large-scale, publicly available datasets, existing segmentation methods based on DL are difficult to apply to this field. In addition, those methods cannot be perfectly applicable because they mainly target irregular objects, whereas airport runway segmentation is for objects with more regular shapes.

\begin{figure*}[hbtp]
\centering
\subfigure[The dataset in \cite{akbar2019runway}]
{
    \begin{minipage}[b]{.31\linewidth}
        \centering
        \includegraphics[scale=0.55]{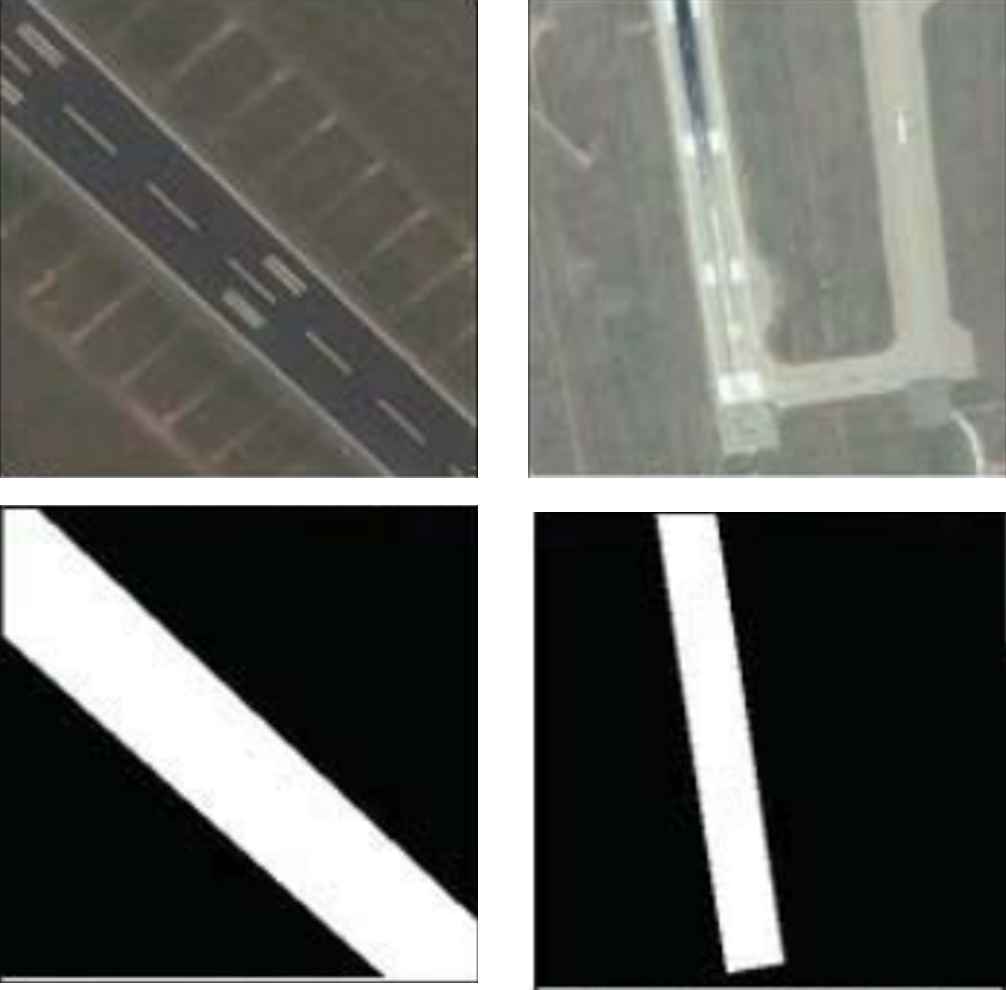}
    \end{minipage}
}
\subfigure[The dataset in \cite{men2020airport}]
{
    \begin{minipage}[b]{.31\linewidth}
        \centering
        \includegraphics[scale=0.55]{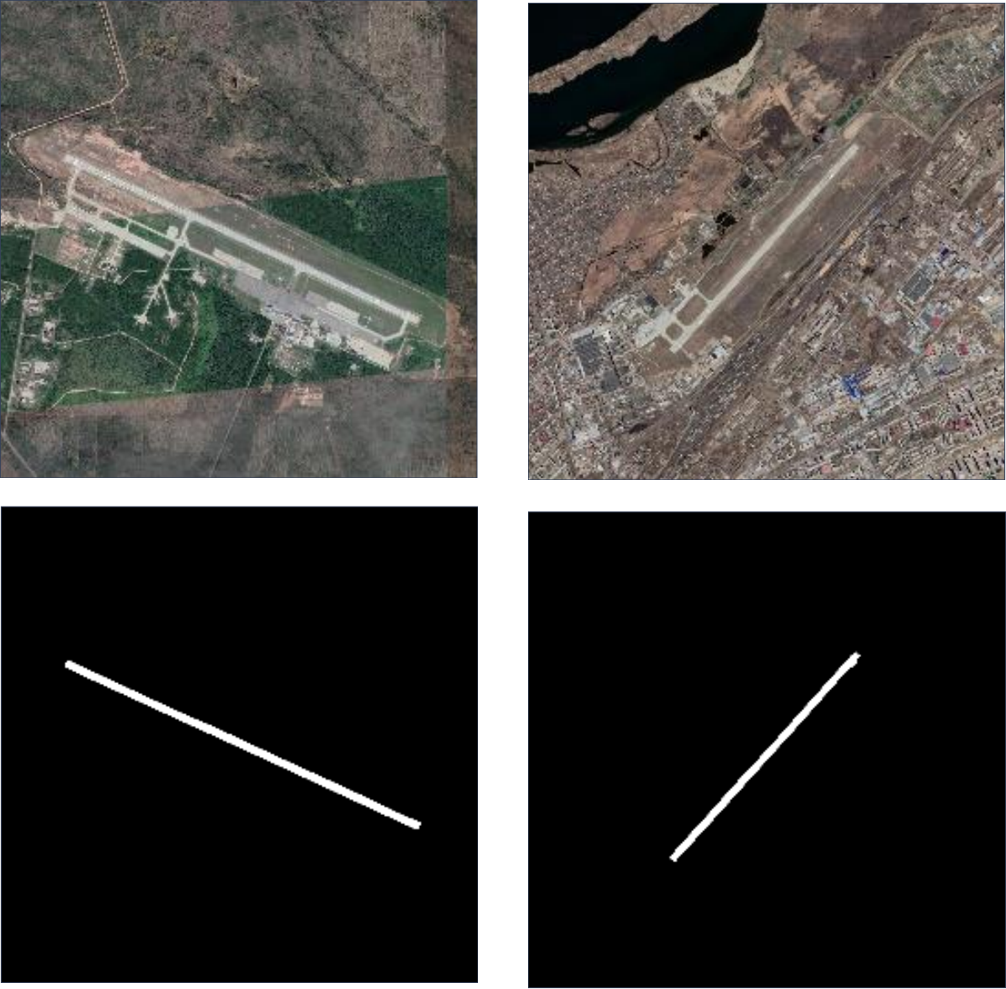}
    \end{minipage}
}
\subfigure[RunwayDataset \cite{wangsemantic}]
{
 	\begin{minipage}[b]{.31\linewidth}
        \centering
        \includegraphics[scale=0.55]{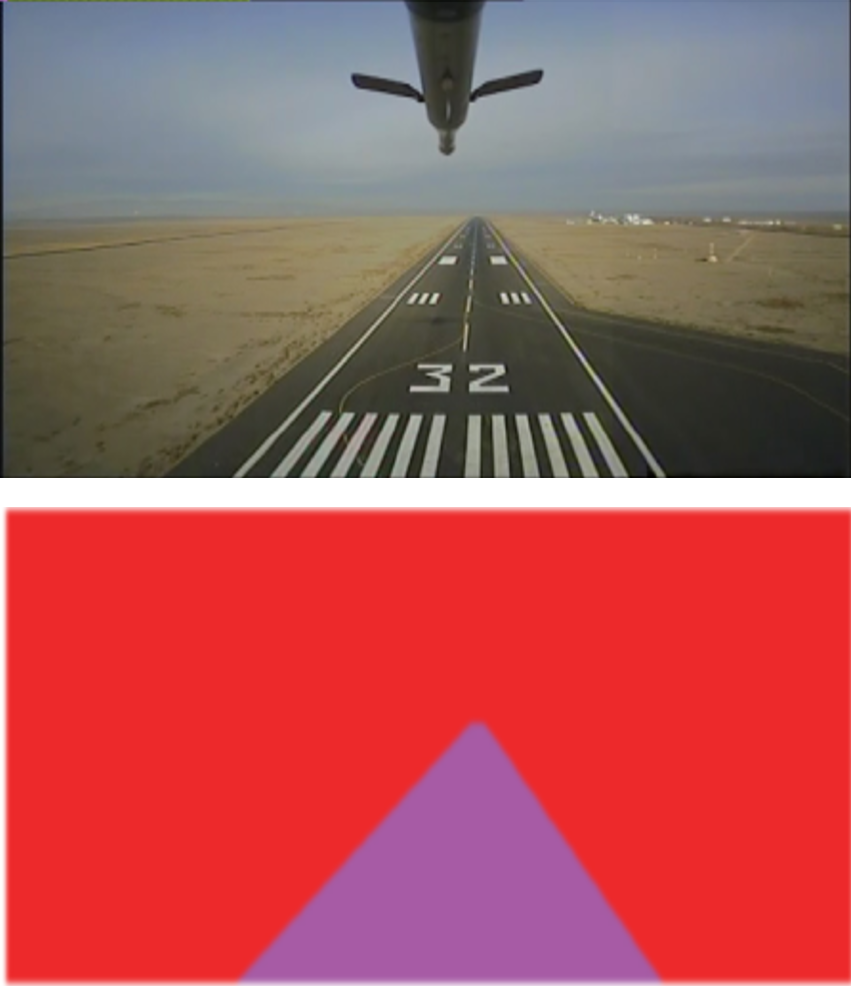}
    \end{minipage}
}
\subfigure[BARS]
{
 	\begin{minipage}[b]{0.98\linewidth}
        \centering
        \includegraphics[width=6in,keepaspectratio]{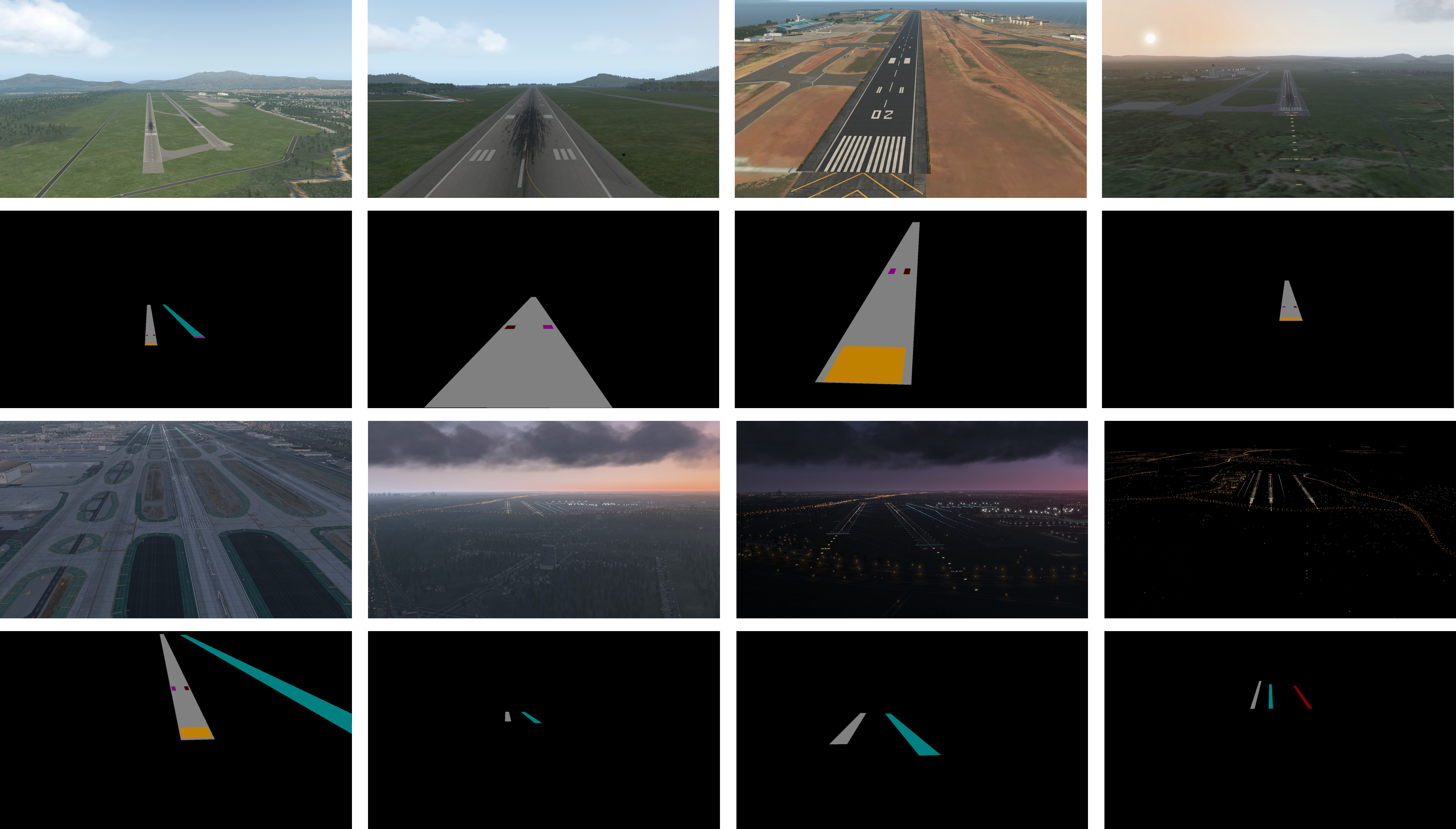}
    \end{minipage}
}
\caption{Some examples from (a) dataset in \cite{akbar2019runway}, (b) dataset in \cite{men2020airport}, (c) RunwayDataset \cite{wangsemantic}, and (d) BARS (ours)}\label{fig-example}
\end{figure*}

To address the two issues raised above, namely, the lack of large-scale, publicly available datasets and the inapplicability of existing methods, we propose a benchmark for airport runway segmentation (BARS), along with a smoothing postprocessing module (SPM) and a contour point constraint loss (CPCL) function. Furthermore, the average smoothness (AS) is designed to measure the smoothness of the segmentation results. BARS contains 10,256 airport runway images from the aircraft view, with images captured from the X-Plane simulation platform $\footnote{X-Plane, \href{https://www.x-plane.com/}{https://www.x-plane.com/}}$. There are 30,201 instances with three categories in BARS. The LabelMe toolbox $\footnote{Labelme, \href{http://labelme.csail.mit.edu/Release3.0/}{http://labelme.csail.mit.edu/Release3.0/}}$ is used to complete the annotation, and a semiautomatic annotation pipeline is designed to reduce the annotation workload.
Compared with other datasets \cite{aytekin2013texture,akbar2019runway,men2020airport,wangsemantic}, the proposed BARS 1) has the largest number of images, 2) contains the fullest categories and instance annotations, 3) involves a variety of scenes, and 4) holds large variations as the images are obtained in different weather and at different times.
Some examples from BARS and other datasets are shown in Fig. \ref{fig-example}.
We employ instance segmentation methods to simultaneously segment different runway markings and to distinguish instances, such as multiple runways within an image.
Eleven representative instance segmentation methods, which include mask-based \cite{maskrcnn,yolact++,mask2former} and contour-based methods \cite{e2ec,liu2021dance}, are evaluated on BARS. SPM and CPCL are proposed based on the regular shape airport runway characteristic. 
SPM is a plug-and-play module that is designed for the inference phase of mask-based instance segmentation methods. SPM employs coarse to fine smoothing operations to alleviate the problem of rough segmentation boundaries. CPCL is proposed for contour-based methods, which can smooth boundaries and speed up the convergence of the model by introducing prior knowledge to restrict the contour points.

The main contributions of this paper are as follows. 

(1) We propose a publicly available benchmark for airport runway segmentation, named BARS. BARS has the largest dataset with the richest categories and the only instance annotation in the field. We also create a semiautomatic annotation pipeline.
Eleven representative instance segmentation methods are evaluated on BARS.

(2) Based on the regular shape airport runway characteristic, we propose SPM, CPCL, and an evaluation metric named AS.

(3) Extensive experiments demonstrate that existing instance segmentation methods can provide prediction results with good performance on BARS. SPM and CPCL can effectively enhance the AS metric while modestly improving accuracy.

\section{Related work}

In this section, we provide a brief overview of existing airport runway segmentation methods and the current progress of instance segmentation.

\subsection{Airport runway segmentation}

Airport runway segmentation methods include traditional image processing-based methods and machine learning-based methods, with traditional image processing-based methods being more common.

\subsubsection{Traditional image processing-based methods}

Traditional image processing-based methods for airport runway segmentation rely on line segments \cite{zhang2022runway} and saliency features \cite{zhang2018airport}. In \cite{budak2016efficient}, the runway was extracted using the line segment detector (LSD) algorithm, but it was insufficient to eliminate boundary information and spurious line segments. Ajith et al. \cite{ajith2019robust} detected the runway boundaries by selecting the appropriate Hough lines using runway characteristics. Abu-Jbara et al. \cite{abu2015robust} proposed a method that combines segmentation and minimization of the energy function.

Traditional image processing-based methods employ artificial features that heavily depend on prior knowledge. Therefore, these methods are effective in some situations, but their application scenarios are quite limited. Unrelated objects having similar shapes or structures, such as rivers, roads, and coastlines, may degrade the performance.

\subsubsection{Machine learning-based methods}

Aytekin et al. \cite{aytekin2013texture} proposed a texture-based method that used the AdaBoost algorithm to segment runways, and a dataset consisting of 57 large satellite images was utilized for the experiment. In \cite{akbar2019runway}, 700 remotely sensed images provided by NWPU-RESISC45 \cite{cheng2017remote} were annotated, and Mask R-CNN was used to accomplish the segmentation. Men et al. \cite{men2020airport} collected 1,300 remote sensing images from "Google Earth" and used DeepLab \cite{chen2017deeplab} to complete runway segmentation. Wang et al. \cite{wangsemantic} proposed the RunwayDataset with 2,000 images for semantic segmentation.

Images from existing runway segmentation datasets can be classified as remote sensing (Earth view) \cite{aytekin2013texture,akbar2019runway,men2020airport} or natural scene (aircraft view) \cite{wangsemantic}. Remote sensing images are taken from the Earth view, which cannot meet the needs of the aircraft landing phase and are not applicable to the visual navigation system. The current datasets for airport runway segmentation are limited in scale and scene, with only one category and no instance annotations. Moreover, these datasets are not publicly available.

\subsection{Instance segmentation}

Instance segmentation is a challenging task because it requires segmenting different categories while distinguishing instances of the same category.

\subsubsection{Mask-based instance segmentation methods}

Mask-based methods segment instances by classifying each pixel. Classical mask-based two-stage instance segmentation methods include bounding box extraction and pixel-level segmentation, such as Mask R-CNN \cite{maskrcnn} and PAnet \cite{wang2019panet}. BMask R-CNN \cite{cheng2020boundary} adds object boundary information to supervised networks to enhance mask prediction. One-stage mask-based methods, such as YOLACT \cite{bolya2019yolact,yolact++}, SOLO \cite{wang2020solo,wang2020solov2}, CenterMask \cite{Centermask}, and CondInst \cite{CondInst}, remove the proposal generation and feature repooling steps, achieving comparable results with higher efficiency.
In recent years, some mask-based methods \cite{solq,queries,swin2,mask2former} inspired  by DETR \cite{DETR} have treated segmentation as a set prediction problem. These methods jointly perform classification, detection, and mask regression on objects of interest using queries as their representation.

\subsubsection{Contour-based instance segmentation methods}

Contour-based methods represent a mask by its contour. In general, contour-based methods outperform mask-based methods in terms of speed. PolarMask \cite{xie2020polarmask} formulates the instance segmentation problem as instance center classification and dense distance regression in polar coordinates. Deep snake \cite{peng2020deep}, DANCE \cite{liu2021dance}, and Point-Set Anchors \cite{wei2020point} deform an initial contour to match the object boundary. E2EC \cite{e2ec} introduces a learnable contour initialization architecture to improve the quality of contour extraction. Contour-based methods have the advantages of easy optimization and fast inference.

Instance segmentation methods are utilized for airport runway segmentation to segment out different categories while distinguishing instances. However, existing advanced methods cannot be perfectly applicable because they mainly target irregular objects, whereas airport runway segmentation is for objects with regular shapes. Therefore, we make improvements based on the characteristics of airport runway markings to alleviate the problem of rough segmentation boundaries.

\section{Dataset}

Datasets play a significant role in data-driven research. The lack of relevant datasets in runway segmentation tasks has become one of the major obstacles. 

\subsection{Image collection}

Obtaining a large quantity of airport runway images from the aircraft view is challenging and necessitates airline cooperation. Thus, we collect images using the FAA-certified X-Plane simulation platform, which is utilized by air forces and aircraft manufacturers for flight instruction. The X-Plane simulation platform encompasses practically all of the world's terrain and is designed according to real-world scenarios. We collect runway images from various airports, aircraft views, weather conditions, and time intervals. These images have been cleaned to ensure availability. For example, exceedingly similar images and images that cannot be annotated due to remote distance are removed.

\subsection{Category selection}

The existing datasets \cite{aytekin2013texture,akbar2019runway,men2020airport,wangsemantic} for airport runway segmentation contain only one category (runway), which is not sufficient for the landing phase. Three categories are carefully selected and annotated in BARS to satisfy practical demands, including runway, threshold marking, and aiming marking.

\begin{figure}[htbp]
\centering
 	\begin{minipage}[b]{1\linewidth}
        \centering
        \includegraphics[width=3.3in,keepaspectratio]{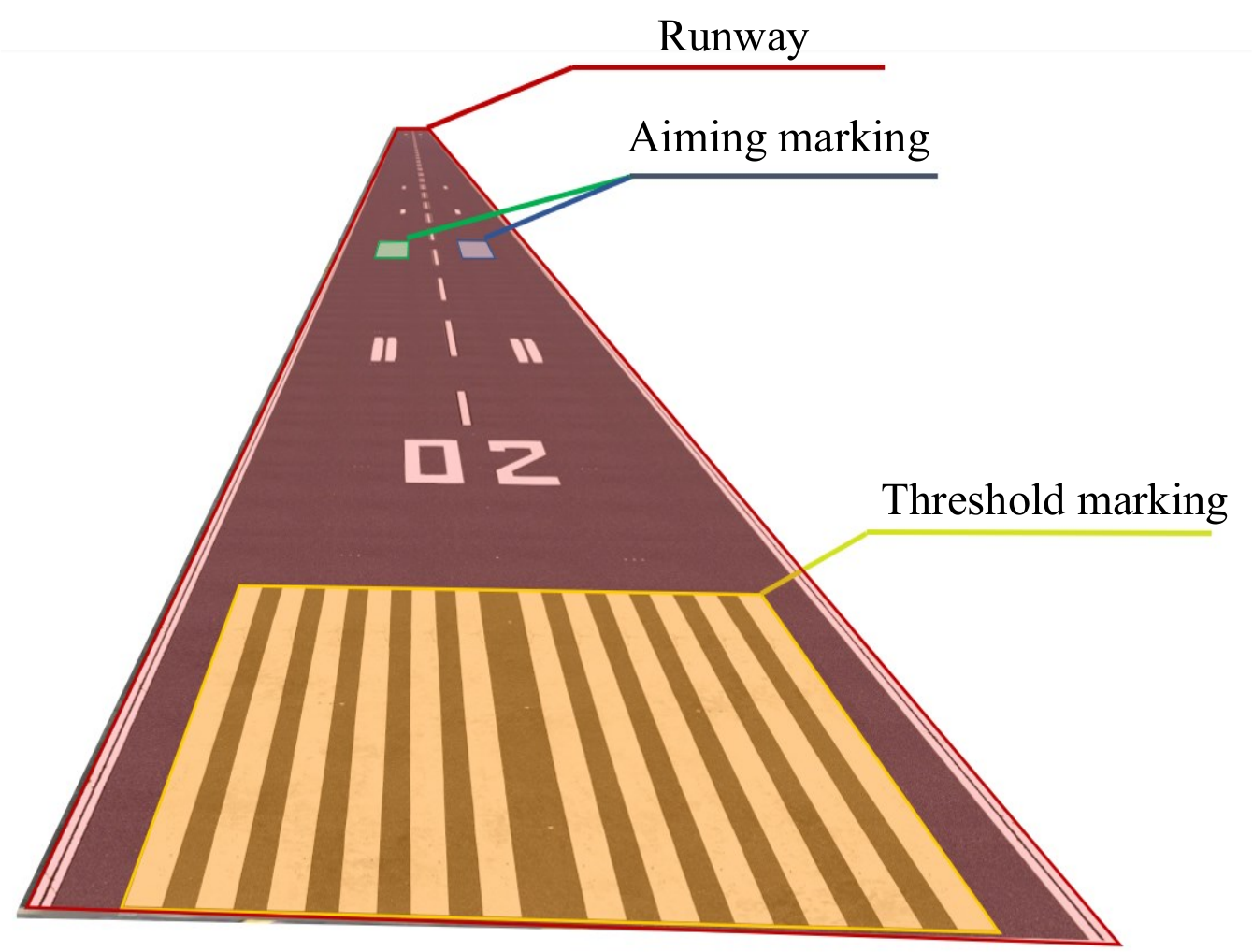}
    \end{minipage}
\caption{Example of the three categories annotated in BARS}\label{Fig_category}
\end{figure}

A runway is a long area for aircraft to takeoff or land, marked by a solid white runway boundary. Threshold marking indicates the entrance to the runway, which consists of a set of longitudinal line segments. Aiming marking provides pilots with the visual location of the landing aiming points. It is made up of two distinct strips, symmetrically located on either side of the runway centerline. For the two strips of the aiming marking, we annotate them as two instances. The specific annotation of the three categories is shown in Fig. \ref{Fig_category}.

\begin{figure*}[!b]
\centering
 	\begin{minipage}[b]{1\linewidth}
        \centering
        \includegraphics[width=6.6in,keepaspectratio]{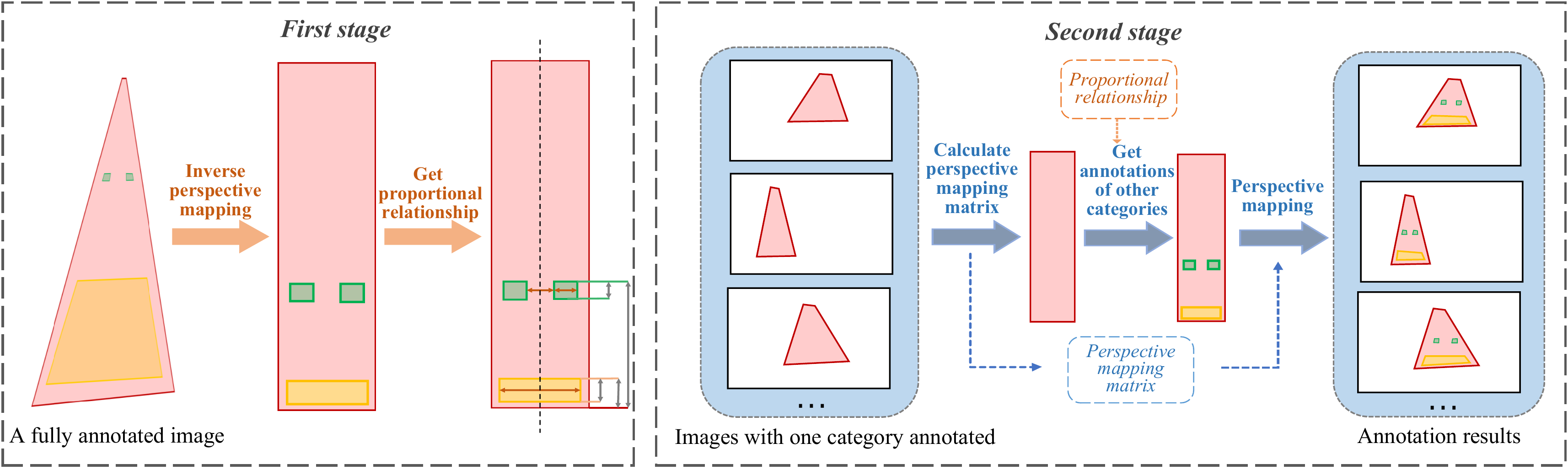}
    \end{minipage}
\caption{The process of semiautomatic annotation pipeline}\label{Fig_semi}
\end{figure*}

There are different types and levels of airport runways, and the markings on each type or level of airport runway differ. For example, there is no threshold marking or aiming marking on some runways. Furthermore, in nighttime scenarios, only the runway is illuminated by runway light. Hence, for runways without threshold markings and aiming markings or in nighttime scenarios, we only annotated the runway category. The selected markings are sufficient for the visual navigation system, so we do not annotate other runway markings, such as the runway number and runway centerline.

\subsection{Semiautomatic annotation pipeline}

We design a semiautomatic annotation pipeline to reduce the annotation workload.
The semiautomatic annotation pipeline allows us to annotate only one category to obtain annotations of other categories. The pipeline is based on the principle that there is a specific proportional relationship between the three categories for the same airport runway. For semiautomatic annotation, the runway images need to be grouped. Since different runways generally have different sizes and proportional relationships, we divide the same runways into a group. For example, images containing runway number 18R of Beijing Capital Airport are grouped together, which often have different views or different weather. Taking one of the groups as an example, the process of the semiautomatic annotation pipeline is shown in Fig. \ref{Fig_semi}.

The pipeline can be divided into two stages: the first stage serves to determine the proportional relationship, which is used in the second stage to obtain the annotation results. As shown in Fig. \ref{Fig_semi}, in the first stage, a single image is selected and annotated with entire categories. The annotated categories are transformed into regular rectangles (shapes in the real-world) by inverse perspective mapping. Specifically, we transform the runway into a rectangle in the top view of the world coordinate system (coordinates of the rectangle can be freely set), calculate the perspective mapping matrix, and then use this matrix to transform the aiming marking and the threshold marking to the top view as well. Finally, the proportional relationship (relative size and position) between the three categories is calculated.

In the second stage, we only annotate one category (e.g., runway) for other images in the group. The image is then converted to top view of the world coordinate system, and the perspective mapping matrix is calculated. At this point, we use the proportional relationship to obtain the annotations of other categories. Finally, perspective mapping is employed to restore the original view of the image.

\subsection{Properties of BARS}

BARS contains 10,256 images and 30,201 instances, with three categories. BARS follows the same format as that in MSCOCO, which can be easily applied to existing instance segmentation methods. The resolution of the images in the dataset is 1920 $\times$ 1080 pixels. Table \ref{tab-data-compasion} shows the differences between BARS and similar datasets.

\begin{table*}[htbp]
\renewcommand{\arraystretch}{1.3}
\caption{Comparison among BARS and other datasets for airport runway segmentation. The proposed BARS provides more categories and images in comparison with other datasets. The images in BARS are taken from the aircraft view and provide instance annotations}\label{tab-data-compasion}
\centering
\begin{tabular*}{\hsize}{@{}@{\extracolsep{\fill}}@{}lcccccr@{}}
\toprule
Datasets & Categories & Images & Instance annotation & Image resolution  & Publicly available & Perspective \\
\midrule 
Dataset in \cite{aytekin2013texture} & 1 & 53 & False & 14000 $\times$ 11000  & False & Earth view\\
\midrule 
Dataset in \cite{akbar2019runway} & 1 & 700 & False & 224 $\times$ 224  & False & Earth view\\
\midrule
Dataset in \cite{men2020airport}& 1 & 1,300 & False & 1536$\times$1536  & False &  Earth view\\
\midrule
Runway dataset \cite{wangsemantic} & 1 & 2,000 & False & 1242 $\times$ 820  & False & Aircraft view  \\
\midrule
BARS (ours) & 3 & 10,256  & True & 1920 $\times$ 1080 & True & Aircraft view \\
\bottomrule
\end{tabular*}
\end{table*}

The advantages of BARS are as follows.

(1) BARS contains 10,256 images and 30,201 instances, which is the largest dataset of the existing runway segmentation datasets.

(2) BARS offers instance annotations and three carefully chosen categories, including runway, threshold marking, and aiming marking, making it the dataset with the fullest categories.

(3) The airports in BARS are diverse, with approximately 40 airports in approximately 15 countries. The majority of them are international airports, such as Zurich Airport, Beijing Daxing Airport, and Los Angeles International Airport.

\begin{figure}[htbp]
\centering
 	\begin{minipage}[b]{1\linewidth}
        \centering
        \includegraphics[width=3.1in,keepaspectratio]{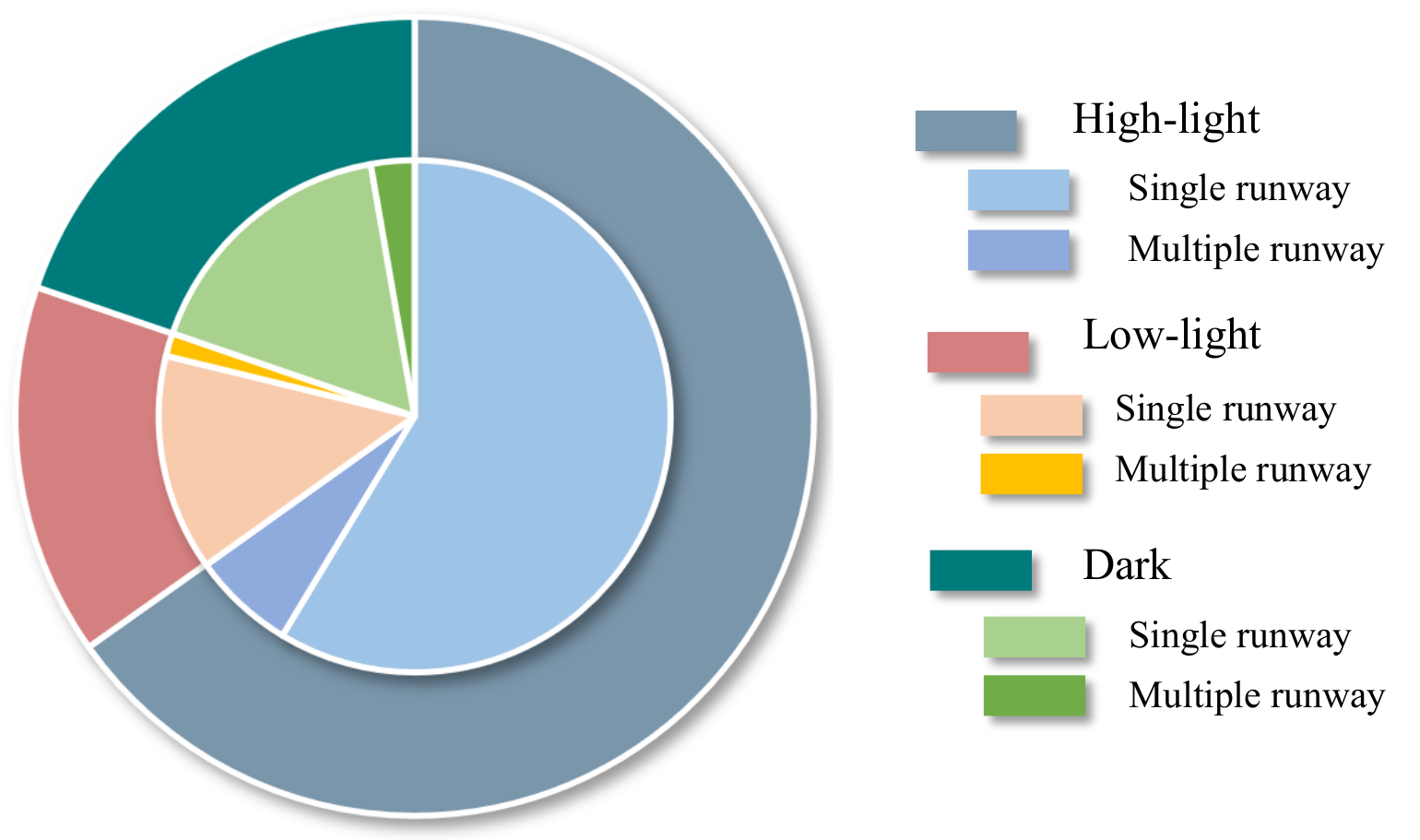}
    \end{minipage}
\caption{Distribution of diverse illumination conditions and different numbers of runways in BARS. The outer circle of the pie chart represents different illumination conditions, which are high-light, low-light, and dark. The inner circle provides the distribution of the number of runways in these conditions, where the single runway is more common}\label{Fig_statisitics}
\end{figure}

(4) Since the images are obtained at different times and in different weather conditions, BARS involves different illumination conditions. Fig. \ref{Fig_statisitics} shows the image number distribution of BARS for diverse illumination conditions and different numbers of runways.

\begin{figure}[hbtp]
\centering
 	\begin{minipage}[b]{1\linewidth}
        \centering
        \includegraphics[width=3.33in,keepaspectratio]{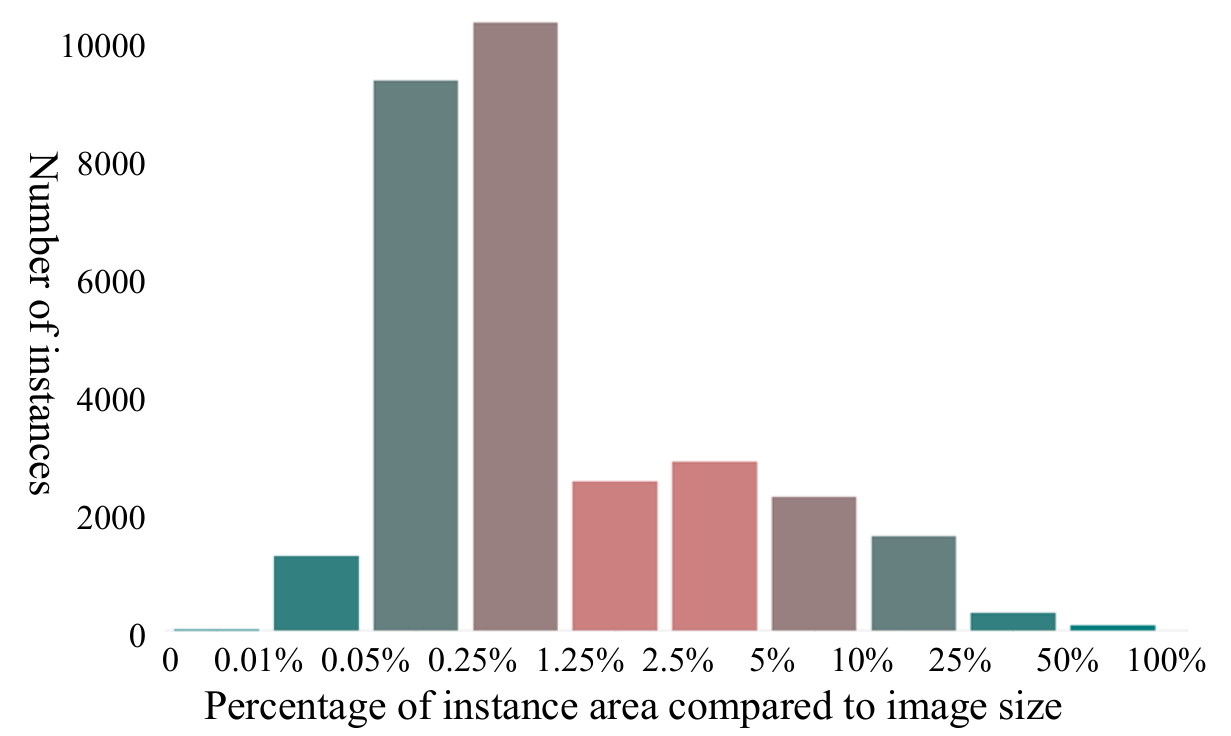}
    \end{minipage}
\caption{Distribution of the number of instances in terms of area size}\label{fig_precentage_area}
\end{figure}

(5) The images in BARS largely simulate the aircraft landing phase, from the aircraft being at an altitude of 500 meters until it arrives on the runway. Therefore, the instances in BARS are distributed in different sizes, as shown in Fig. \ref{fig_precentage_area}. It should be noted that aiming markings are frequently small objects, making instance segmentation on BARS challenging.

In summary, we propose a dataset specifically for airport runway segmentation. This dataset focuses on the aircraft landing phase and can meet the needs of practical applications. Our work fills a gap in the missing airport runway segmentation dataset.

\section{Method}

We trained and evaluated eleven representative instance segmentation methods on BARS. Based on the regular shape airport runway characteristic, we propose SPM and CPCL to improve existing methods. Among them, SPM is proposed for the inference phase of mask-based methods, while CPCL is proposed for the training phase of contour-based methods.

\begin{figure*}[htbp]
\centering
 	\begin{minipage}[b]{1\linewidth}
        \centering
        \includegraphics[width=6.3in,keepaspectratio]{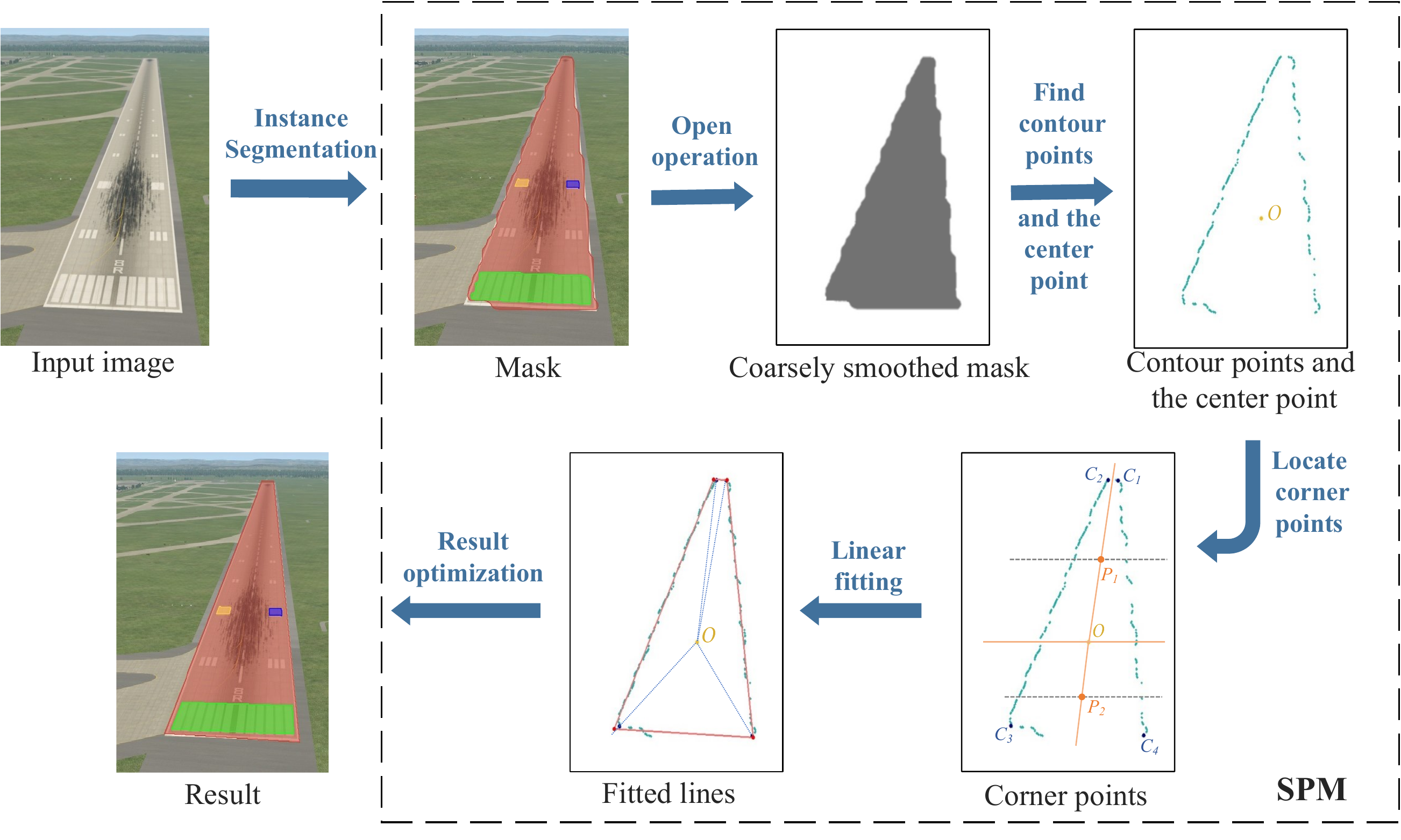}
    \end{minipage}
\caption{Flow chart for SPM}\label{fig-SPM}  
\end{figure*}

\subsection{Main idea of SPM}

We find that mask-based instance segmentation methods have rough boundaries due to pixel-by-pixel classification. In particular, the two-stage methods, while more precise, do not perform well on boundary information for large objects. Therefore, we designed the SPM to optimize the boundary problem by fitting four lines.

SPM is utilized for instances where the size is larger than 0.1\% of the image size, as it may cause a loss of accuracy in instances where the size is too small.
This module requires no training and can be applied to models at the inference phase with ease. As shown in Fig. \ref{fig-SPM}, SPM consists of the following steps: 

(1) Open operation. In image morphology, the formula for the open operation is expressed as:

\begin{equation}\label{0101}
B \circ X = (B \ominus X)\oplus B
\end{equation}

where $\ominus$ is the erosion operation, $\oplus$ is the dilation operation, $X$ is the mask of the model output, and $B$ is the basic morphological operator.

The size of $B$ can be set according to the characteristics of the mask obtained from the different methods. For example, we set the size of $B$ to $1\times1$ for Mask2Former. The larger the size of $B$ is, the greater the degree of smoothing, but it will affect mask accuracy to some extent. The open operation can coarsely smooth the boundary while preserving the original shape of the mask. We repeat the open operation three times for each mask to obtain the coarsely smoothed mask.

(2) Find contour points and the center point. The contour tracking algorithm \cite{suzuki1985topological} is used to obtain the contour points of the coarsely smoothed mask. The number of contour points obtained is enormous since the coarsely smoothed mask is not totally regular.Moment, a specific quantitative measurement of the shape of a set of points, is used to locate the center point. Specifically, the center point  is calculated by dividing the first moment by the zeroth moment, which is denoted as $O$.

(3) Locate the four corner points. Make transversal lines above and below the center point. Each transversal line intersects the mask twice, and the intersection's center point serves as the positioning point. Donate two positioning points as $P_{1}$ and $P_{2}$. As shown in Fig. \ref{fig-SPM}, the positioning point and the center point divide each contour point into four sets. The corner point, denoted by $C_{i}(i \in [1,2,3,4])$, is determined as the point in each set that is farthest from the center point.

(4) Linear fitting. Use the angular interval between the four corner points and the center point to divide all contour points into four point sets. Perform least squares linear fitting for each point set separately to obtain four linear equations, corresponding to the four edges of the instance. Enclose the quadrilateral by the four lines to yield the result. Instead of fitting the quadrilateral directly to the four corner points, all contour points were used to ensure the accuracy of the results.

(5) Result optimization. To eliminate bad cases, filter the results. If the Intersection-over-Union (IoU) of the result to the original mask is less than the set threshold, we consider that an error has occurred. In this case, use the result of polygon approximation for the contour points as the final fine result. Specifically, the Douglas-Peucker algorithm \cite{douglas1973algorithms} is used to process contour points. A maximum tolerance distance is required by the algorithm; the greater the maximum tolerance distance, the fewer key points will be obtained. Set the maximum tolerance distance to be adaptive depending on the size of the instance, which is defined as equation \ref{distance}.

\begin{equation}\label{distance}
d = 0.01 \times log(length)
\end{equation}

where $length$ denotes the perimeter of the shape formed by the contour points.

In summary, by smoothing the mask from coarse to fine directly in the inference phase, SPM can improve the mask boundary while maintaining the original shape structure.

\subsection{Main idea of CPCL}

Contour-based instance segmentation methods usually fit the mask with a fixed number of contour points. For example, 128 points are typically used in E2EC and 36 points in PolarMask. However, airport runway markings have a regular shape, so only a few corner points are required to fit them. The existing loss functions focus on the correctness of contour points rather than the smoothness of the boundary fitted by contour points. This leads to the fact that existing methods may fit a rough shape with slightly more contour points. Therefore, we designed the CPCL.

The CPCL can be divided into two steps. First, the fixed number of instance contour points predicted by the network is processed using the Douglas-Peucker algorithm to obtain the key points. The more key points we obtain, the rougher the shape of the contour is. Second, $smoothL_{1}$ loss is used for supervision after obtaining the number of key points. The CPCL is defined as:

\begin{equation}\label{cpcl}
L_{CPC} = smoothL_{1}(\widetilde{x}-x^{gt})
\end{equation}

where $\widetilde{x}$ is the number of key points and $x^{gt}$ is the number of corner points in our annotated polygons.

In summary, CPCL indirectly smooths the segmentation result by restricting the contour points to yield fewer key points in the training phase. In addition, CPCL introduces prior knowledge that the shape of the runway is regular, which accelerates the convergence of the network.

\subsection{Applicability analysis}

In this section, we will discuss the applicability of our proposed methods, namely, why SPM is only used for mask-based methods and CPCL is only used for contour-based methods.

Contour-based methods fit the mask with contour points. The number of these points is quite modest in comparison to mask-based methods that use pixel-by-pixel segmentation, even though they are sufficient to fit the ground truth. Additionally, there are inevitably large deviation contour points in some segmentation results. Therefore, SPM is unsuitable for contour-based methods because it is easily affected by large deviation points in this case, resulting in segmentation results with large errors.

CPCL is intended for contour-based methods and achieves smoothing by restricting contour points. Hence, CPCL cannot be applied to mask-based methods since there are no predefined contour points.

\section{Experiments}

\subsection{Experimental settings}

\subsubsection{Evaluation criteria}

We employed the widely used segmentation evaluation criteria, average precision (AP), provided by the MSCOCO dataset as our evaluation criteria. It first calculated the recall and precision using the following equation \ref{precision} and equation \ref{recall}. The acronyms $TN$, $FN$, $TP$, and $FP$ stand for true-negative, false-negative, true-positive, and false-positive, respectively. The Precision-Recall (P-R) curve was subsequently constructed using the aforementioned precision and recall. The AP evaluation criteria in MSCOCO is the area bounded by the P-R curve and X-Axis, which is calculated as shown in equation \ref{AP}.

\begin{equation}\label{precision}
precision = \frac{TP}{TP+FP}
\end{equation}

\begin{equation}\label{recall}
recall = \frac{TP}{TP+FN}
\end{equation} 

\begin{equation}\label{AP}
AP = \int_{0}^{1}p(r)dr
\end{equation}

The MSCOCO evaluation criteria include six critical metrics: $AP$, $AP_{50}$, $AP_{75}$, $AP_{S}$, $AP_{M}$ and $AP_{L}$. $AP$, $AP_{50}$, and $AP_{75}$ refer to the mean AP of all categories under IoU thresholds of 0.5 to 0.9, 0.5, and 0.75, respectively. $AP_{S}$, $AP_{M}$ and $AP_{L}$ are used to further measure the performance of the methods on segmenting instances of various sizes. The subscripts $S$, $M$, and $L$ denote an area of less than $32 \times 32$, between $32 \times 32$ and $92 \times 92$, and larger than $92 \times 92$, respectively.

We noticed that the existing evaluation metrics focus more on accuracy than smoothness. The smoothness metric should be considered when segmenting regular objects, such as airport runways. Therefore, existing metrics cannot adequately measure the performance of the runway segmentation task. Take the widely used AP metric as an example. The underlying measure used in AP to compare predictions and ground truth is IoU. IoU divides the intersection area of two masks by their union area. This metric assigns equal weight to all pixels and is hence less sensitive to boundary in larger objects, which largely determines the smoothness of the mask.

Based on the above insights, we designed AS to measure the smoothness of the segmentation results. For the masks obtained by different methods, we first obtain the contour points using the contour tracking algorithm \cite{suzuki1985topological}. Then, to obtain refined contour points, we utilize the Douglas-Peucker algorithm to remove the points on the same straight line by setting the maximum tolerance distance to a small value (we take 1). The formula for AS is as follows:

\begin{equation}\label{AS}
\begin{aligned}
AS = x / \log_{2}{(length)}
\end{aligned}
\end{equation}

where $x$ denotes the number of refined contour points, and $length$ represents the perimeter of the shape formed by these points. The smaller the AS is, the smoother the boundary of the mask.

It is worth noting that although the method of obtaining AS is similar to that of SPM, the parameters and purpose are different. In SPM, the maximum tolerance distance is set to an adapted value to make the algorithm approximate the rough boundary as a straight line, while in AS, it is set to a small value to exclude points on the same line so that the smoothness of the mask can be more precisely measured.

In summary, AS can evaluate the smoothness of the segmentation results, which are generally ignored by IoU-based evaluation metrics. We hope that the adoption of the new evaluation will help make faster progress in the task of regular object segmentation.

\begin{table*}[htbp]
\begin{center}
\caption{Results obtained on BARS test set}\label{tab_eleven}
\begin{tabular*}{\hsize}{@{}@{\extracolsep{\fill}}lllllllll@{}}
\toprule
 & Method 
& $AP$ &$AP_{50}$ & $AP_{75}$ & $AP_{S}$ & $AP_{M}$ & $AP_{L}$ & $AS$ \\
\midrule
\multirow{8}*{Mask-base} 
& Mask R-CNN \cite{maskrcnn} 
& 71.17  & 83.69  & 79.52  &  60.07 & 91.28 & 89.59 & 2.72 \\
& BMask R-CNN \cite{cheng2020boundary} 
& 72.03 & 83.75  & 80.29 & 65.52  & 91.59  & 91.36 & 2.98 \\
& Mask2Former \cite{mask2former} 
 & \textbf{77.42} & \textbf{90.98}  & \textbf{83.61} & \textbf{67.82}  & \textbf{95.06}  & \textbf{98.38} & \textbf{1.01} \\

& CondInst \cite{CondInst} 
 & 50.32 & 70.32  & 54.42 & 22.10  & 77.39  & 91.87 & 3.53 \\
 
& YOLACT \cite{bolya2019yolact} 
& 49.47 & 76.97  & 52.50 & 14.16 & 69.81 & 87.12 & 2.86 \\
& YOLACT++ \cite{yolact++} 
& 55.96   & 79.57  & 60.12 & 15.31  & 71.16  & 92.02 & 2.77 \\
& SOLO \cite{wang2020solo} 
 & 43.72 & 62.18 & 48.44 & 20.65 & 61.39 & 76.68  & 3.41 \\
& SOLOv2 \cite{wang2020solov2}
& 48.07 & 65.48  & 52.42 & 24.88 & 67.26 & 86.45  & 4.40\\

\midrule
\multirow{3}*{Contour-base} &
PolarMask \cite{xie2020polarmask}
& 31.70   & 65.51 & 27.95 & 20.88  & 50.74  & 33.21 & 1.58 \\
& DANCE \cite{liu2021dance}
& 72.24 & 86.31 & 79.31 & 58.11 & 90.39 & 88.32 & 1.38\\
& E2EC \cite{e2ec}
& 64.08 & 77.41 & 70.73 & 55.80 & 87.94 & 52.78 & 1.54 \\
\bottomrule
\end{tabular*}\label{tab2}
\end{center}
\end{table*}

\begin{figure*}[htbp]
\centering
 	\begin{minipage}[b]{1\linewidth}
        \centering
        \includegraphics[width=6.2in,keepaspectratio]{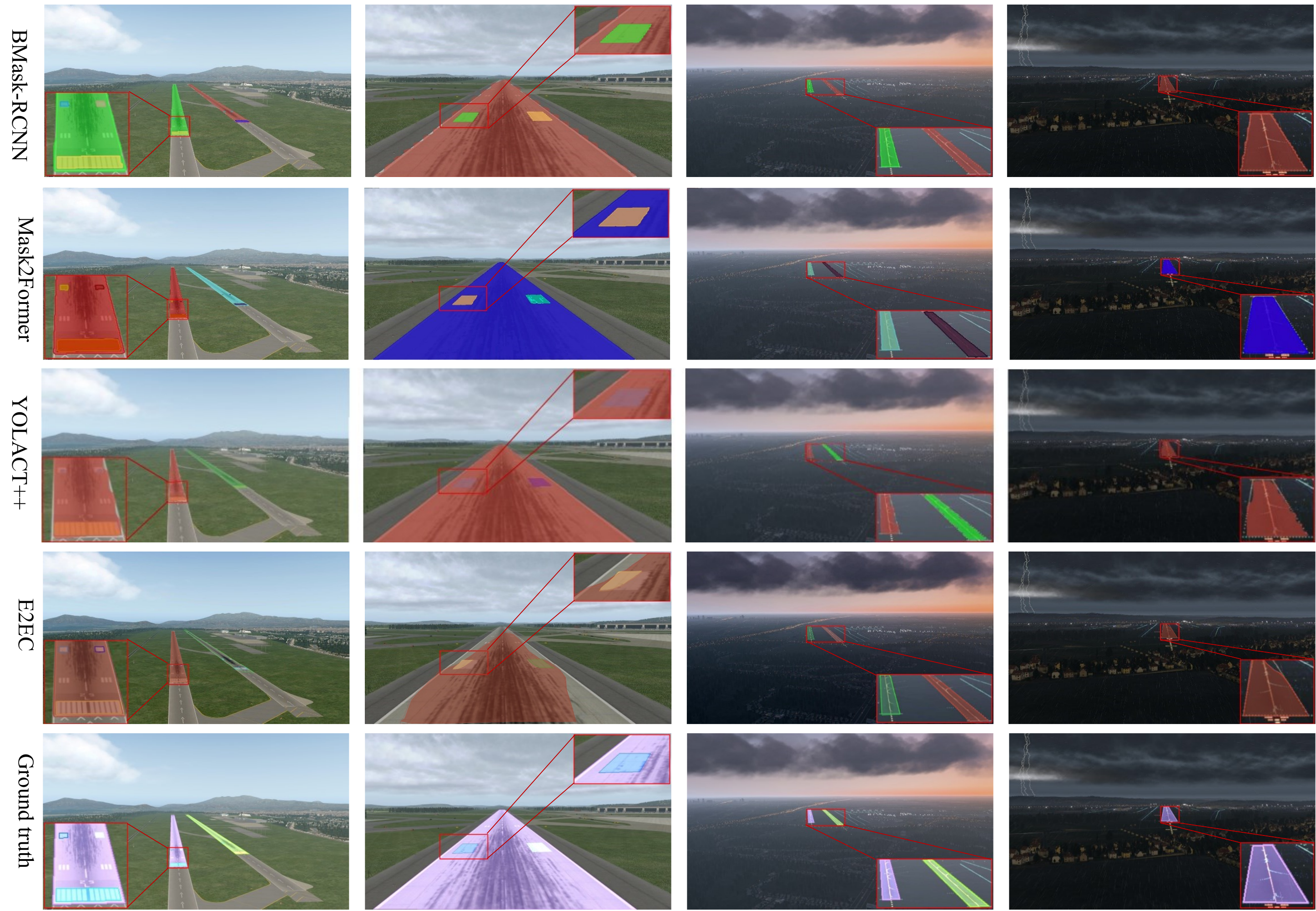}
    \end{minipage}
\caption{BARS example results from BMask R-CNN \cite{cheng2020boundary}, Mask2Former \cite{mask2former}, YOLACT++ \cite{yolact++},  E2EC \cite{e2ec}, and ground truth}\label{fig_vis}
\end{figure*}

\subsubsection{Implementation details}

The BARS dataset is divided into 8,004 training images, 1,002 validation images, and 1,250 testing images. The results of the testing images are taken as the final results.

All of our experiments are carried out using PyTorch. We used ResNet50 with FPN as the backbone except for Mask2Former, which used Swin-L \cite{swin}. The backbones used are pretrained on ImageNet. The image size as well as the optimizer settings are modified to default settings.

\subsection{Evaluation of representative methods}

A total of eleven representative instance segmentation methods are selected for evaluation on our dataset. The quantitative results are shown in Table \ref{tab_eleven}.

We can conclude that (1) the two-stage mask-based method, such as Mask R-CNN and BMask R-CNN, has superior accuracy, particularly for large objects. (2) Mask2Former achieves state-of-the-art results in each of the metrics. (3) YOLACT and SOLO, those one-stage mask-based methods, perform poorly in segmenting small objects, resulting in average overall accuracy. (4) The accuracy of contour-based methods depends on the fitting effect of contour points. Such methods are generally not particularly effective for large objects, especially PolarMask, which uses fixed-angle polar coordinates to represent the mask and may not accurately capture the corner points of objects. (5) Contour-based methods basically outperform mask-based methods in terms of the AS metric. This corresponds to the characteristic of contour-based methods for fitting masks with finite contour points, which also verifies the reasonableness of the metric we designed.

As shown in Fig. \ref{fig_vis}, BMask R-CNN, Mask2Former, YOLACT++, and E2EC are selected to demonstrate the visualization results. The results indicate that (1) BMask R-CNN and Mask2Former are well suited to instances of all sizes. However, BMask R-CNN does not perform well for boundary information of large objects. (2) YOLACT++ has poor segmentation results for instances of smaller size. (3) E2EC does not perform robustly. If the corner points are well fitted, the effect can be impressive; otherwise, the results will be much worse than mask-based methods.

\subsection{The effect of SPM}

\begin{figure*}[htbp]
\centering
 	\begin{minipage}[b]{1\linewidth}
        \centering
        \includegraphics[width=6.4in,keepaspectratio]{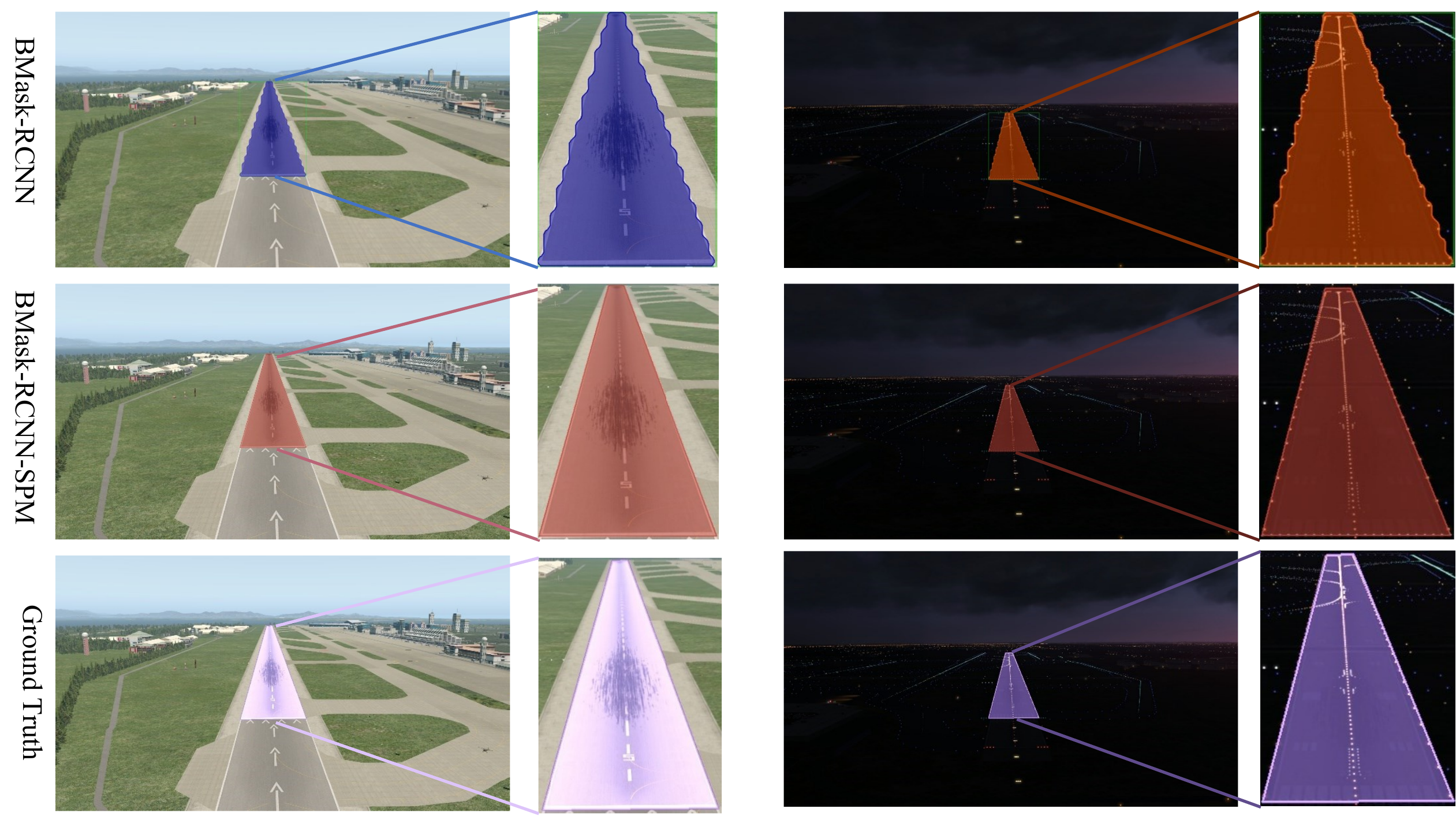}
    \end{minipage}
\caption{Qualitative results of SPM. Some prediction examples of BMask R-CNN, BMask R-CNN with SPM, and ground truth}\label{pic_pp}
\end{figure*}

Experiments to evaluate the SPM effect were conducted for BMask R-CNN, Mask2Former and YOLACT. The results are shown in Table \ref{tab_pp}. The experimental results demonstrate that SPM can be applied to various methods and achieve a modest accuracy improvement. In addition, the AS metric is significantly improved by 30\% for methods such as BMask R-CNN and YOLACT, where the output is coarser.
Fig. \ref{pic_pp} shows the visualization results of SPM applied to BMask R-CNN, which shows that SPM can smooth prediction results well and obtain accurate and refined boundaries.

\begin{table*}[htbp] 
\renewcommand{\arraystretch}{1.3}
\begin{center}
\caption{Results of SPM combined with existing methods}\label{tab_pp}%
\begin{tabular*}{\hsize}{@{}@{\extracolsep{\fill}}l lll lll l@{}}
\toprule
Method & $AP$ &$AP_{50}$ & $AP_{75}$ & $AP_{S}$ & $AP_{M}$ & $AP_{L}$ & $AS$\\
\midrule
BMask R-CNN
& 72.03 & \textbf{83.75}  & \textbf{80.29} & \textbf{65.52}  & \textbf{91.59}  & 91.36 & 2.98 \\
BMask R-CNN-SPM
& \textbf{72.13} & 83.73  & 80.28 & \textbf{65.52}  & 91.58  & \textbf{91.96} & \textbf{1.79} \\
\midrule
Mask2Former
 & 77.42 & \textbf{90.98}  & 83.61 & \textbf{67.82}  & 95.06  & \textbf{98.38} & 1.01 \\
 Mask2Former-SPM
 & \textbf{77.43} & \textbf{90.98}  & \textbf{83.64} & \textbf{67.82}  & \textbf{95.11}  & 98.36 & \textbf{0.96} \\
\midrule
 YOLACT
& 49.47 & \textbf{76.97}  & \textbf{52.50} & 14.16 & \textbf{69.81} & 87.12 & 2.86 \\
 YOLACT-SPM
& \textbf{49.52} & 76.96  & \textbf{52.50} & \textbf{14.27} & 69.47 & \textbf{87.53} & \textbf{1.65} \\
\bottomrule
\end{tabular*}
\end{center}
\end{table*}

\subsection{The effect of CPCL}

\begin{table*}[htbp]
\renewcommand{\arraystretch}{1.3}
\begin{center}
\caption{Results of CPCL combined with existing methods}\label{tab_cpcl}
\begin{tabular*}{\hsize}{@{}@{\extracolsep{\fill}}l lll lll l@{}}
\toprule
Method & $AP$ &$AP_{50}$ & $AP_{75}$ & $AP_{S}$ & $AP_{M}$ & $AP_{L}$ & $AS$  \\
\midrule
PolarMask & 31.70  & 65.51 & 27.95 & \textbf{20.88}  & 50.74  & \textbf{33.21} & 1.58 \\
PolarMask-CPCL & \textbf{32.01}  & \textbf{66.73} & \textbf{28.34} & 20.66  & \textbf{51.53}  & 29.35  & \textbf{1.22} \\
\midrule
DANCE 
& 72.24 & 86.31 & 79.31 & 58.11 & 90.39 & 88.32 & 1.38 \\
DANCE-CPCL
& \textbf{74.77} & \textbf{86.45} & \textbf{81.10} & \textbf{61.00} & \textbf{93.11} & \textbf{91.22} & \textbf{1.27}\\
\midrule
E2EC 
& 64.08 & 77.41 & \textbf{70.73} & \textbf{55.80} & 87.94 & \textbf{52.78} & 1.54 \\
E2EC-CPCL
& \textbf{64.16} & \textbf{79.53} & 70.08 & 47.89 & \textbf{88.29} & 52.57 & \textbf{1.46} \\
\bottomrule
\end{tabular*}
\end{center}
\end{table*}

We evaluate the effect of CPCL using PolarMask, DANCE and E2EC as baselines, with all hyperparameters kept the same. Table \ref{tab_cpcl} shows that our proposed loss function can effectively improve the AP and AS metrics. For example, CPCL improves DANCE by approximately 3\% and 8\% in terms of the AP metric and AS metric, respectively.

\begin{figure}[htbp]
\centering
 	\begin{minipage}[htbp]{1\linewidth}
        \centering
        \includegraphics[scale=0.6]{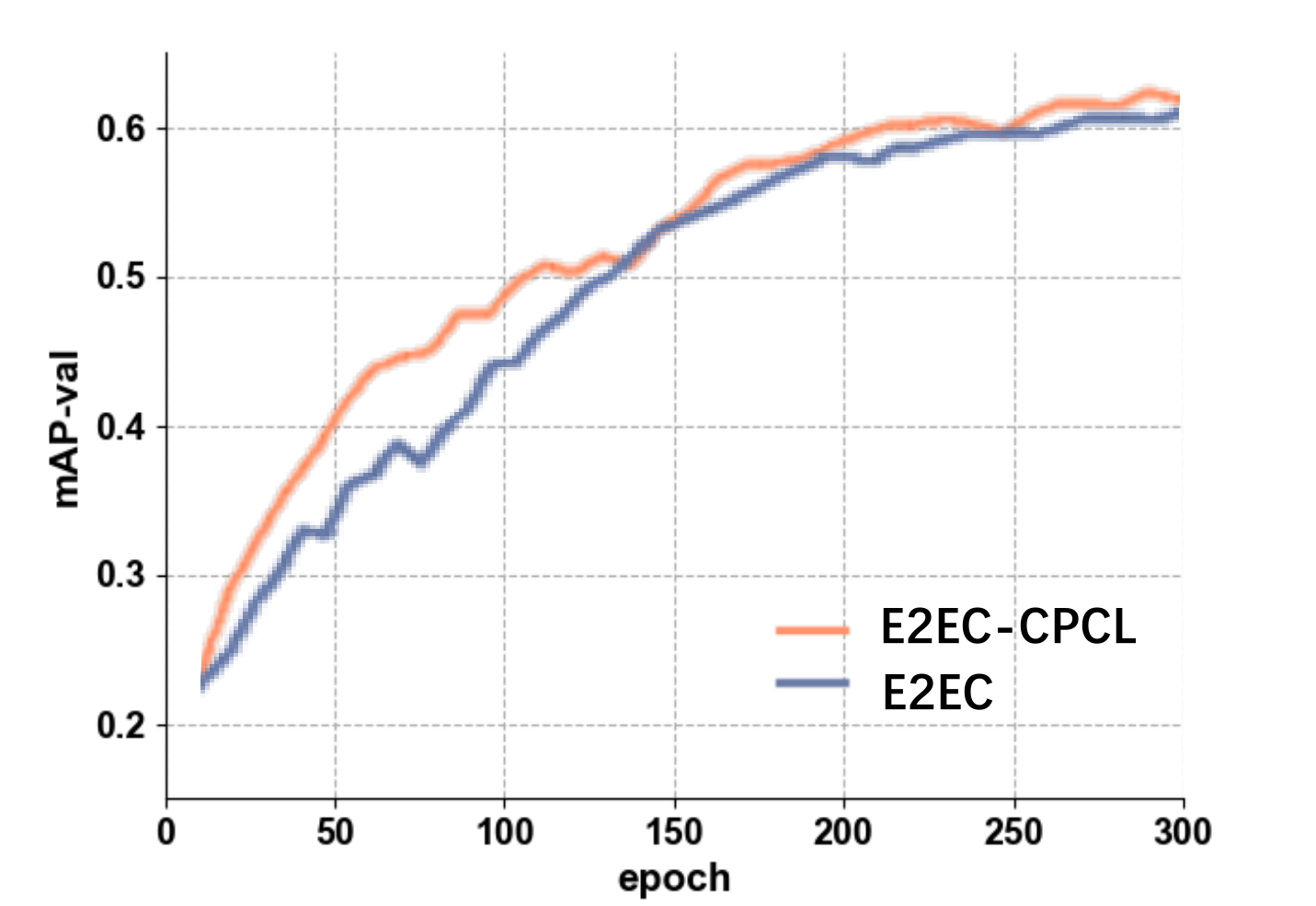}
    \end{minipage}
\caption{The performance of the AP metric on the validation set during the training process. The orange curve in the figure shows the trend of E2EC with CPCL, and the blue curve represents E2EC. Both curves are smoothed by the TensorBoard toolbox}\label{vis_loss}
\end{figure}

Furthermore, CPCL can accelerate the convergence of the network due to the introduction of a priori knowledge that the runway is a regular shape. This is demonstrated in Fig. \ref{vis_loss}, which shows the AP results we recorded for the validation set of E2EC during training.

\subsection{Experiments on application analysis}

From the analysis in Section 4.3, it can be concluded that CPCL cannot be used for mask-based methods and SPM may not be suitable for contour-based methods. Two contour-based methods, DANCE and E2EC, are selected to evaluate the effects of employing SPM alone and SPM and CPCL together. The results are shown in the Table \ref{tab_Applicability}. Despite the fact that contour-based methods can utilize SPM to enhance smoothness, experimental results demonstrate that it leads to a significant loss in accuracy.

\begin{table*}[htbp]
\renewcommand{\arraystretch}{1.3}
\begin{center}
\caption{Results of SPM alone and SPM and CPCL used together in contour-based methods}\label{tab_Applicability}
\begin{tabular*}{\hsize}{@{}@{\extracolsep{\fill}}l lll lll l@{}}
\toprule
Method & $AP$ &$AP_{50}$ & $AP_{75}$ & $AP_{S}$ & $AP_{M}$ & $AP_{L}$ & $AS$  \\
\midrule
DANCE 
& 72.24 & 86.31 & 79.31 & 58.11 & 90.39 & 88.32 & 1.38 \\
DANCE-SPM
& 70.63 & 85.89 & 77.40 & 58.11 & 89.73 & 83.14 & 1.33 \\
DANCE-CPCL
& \textbf{74.77} & \textbf{86.45} & \textbf{81.10} & \textbf{61.00} & \textbf{93.11} & \textbf{91.22} & 1.27 \\
DANCE-CPCL-SPM
& 72.80 & 85.69 & 78.60 & \textbf{61.00} & 92.59 & 84.53 & \textbf{1.18} \\
\midrule
E2EC 
& 64.08 & 77.41 & \textbf{70.73} & \textbf{55.80} & 87.94 & \textbf{52.78} & 1.54 \\
E2EC-SPM
& 63.01 & 76.75 & 69.50 & 55.76 & 87.52 & 48.45 & 1.49 \\
E2EC-CPCL
& \textbf{64.16} & \textbf{79.53} & 70.08 & 47.89 & \textbf{88.29} & 52.57 & 1.46 \\
E2EC-CPCL-SPM
& 62.52 & 78.58 & 68.15 & 47.89 & 88.09 & 46.71 & \textbf{1.41} \\
\bottomrule
\end{tabular*}
\end{center}
\end{table*}

\section{Limitations}

Our research still has some limitations.

(1)	Even though BARS contains a variety of scenarios, such as night, cloudy, and overcast, some scenarios, like foggy or hazy days, are not included due to the simulation platform's limitations.

(2) SPM can moderately improve accuracy with enhanced smoothness. However, SPM does not provide learnable parameters and is susceptible to low-quality points. In this case, SPM is a trade-off, and accuracy may be sacrificed if smoother segmentation results are desired.

In the future, we will include more scenarios for BARS. Additionally, we will consider converting SPM into a learnable module and introducing the attention mechanism to make it achieve better performance.

\section{Conclusion}

This paper presents a publicly available benchmark for airport runway segmentation, named BARS. BARS has the largest, fullest categories and the only instance annotated dataset in the field. A semiautomatic annotation pipeline is designed to reduce the annotation workload. We evaluate eleven instance segmentation methods on BARS and propose SPM and CPCL. Furthermore, an evaluation metric named AS is designed for measuring smoothness. Extensive experiments show that existing instance segmentation methods have good accuracy on BARS. In addition, the experiments indicate the effectiveness of our proposed SPM and CPCL for this task, improving both accuracy and smoothness. Airport runway segmentation is particularly meaningful and challenging for the aviation industry. We believe that BARS will promote the development of a visual navigation system for aircraft.

\bibliography{References}

\end{sloppypar}
\end{document}